\documentclass{article}
\usepackage{iclr2027_conference}
\usepackage{bda_style}
\title{On numerical issues of divergences between nearly equal distributions}
\author{Brandon Amos \\
Reflection AI \\
\texttt{brandon@reflection.ai}}

\hypersetup{pdfauthor={Brandon Amos}}

\begin{document}
\maketitle
\begin{abstract}
\citet{barron2026softmax} recently identified numerical issues in the
softmax and soft cross-entropy loss. We show that similar issues arise in
divergences between categorical distributions.
This includes the forward and reverse Kullback--Leibler (KL) divergence and
generalized Jensen--Shannon divergence (JSD), as well as Monte Carlo KL
estimates and entropy. When the distributions are nearly equal, rounding
and cancellation issues affect the forward and gradient computations.
These issues can affect settings using divergences in optimization
objectives, such as knowledge distillation, reinforcement learning, and
proximal methods. We propose numerically stable reformulations and provide
example implementations for each setting, validated by property-based tests.
\end{abstract}

\section{Introduction}
\label{sec:introduction}
Standard implementations of the softmax and soft cross-entropy loss subtract the
maximum logit to avoid overflow, but their gradients can still suffer
from rounding and cancellation.
\citet{barron2026softmax} analyze and fix two issues here.
First, \emph{softmax collapse}, previously identified by
\citet{prieto2025grokking}, occurs for confident hard-target predictions.
The target probability rounds to one, making its cross-entropy gradient
zero even though the other categories retain positive probabilities and
nonzero gradients. Retaining the small probability tail avoids this
failure. Second, \emph{softmax cancellation} affects soft cross-entropy
when the prediction and target distributions are nearly equal.
The gradient depends on small differences between their probabilities,
which separate normalization and rounding can erase.
Computed target probabilities need not sum exactly to one, so differentiation
can even produce a nonzero gradient for identical logits. Reconstructing
probability differences from logit differences, together with a zero-sum
correction, addresses this second failure.

These results motivate further studying the numerical accuracy of other
divergences in machine learning settings.
The forward KL inherits the soft cross-entropy gradient issue because
their gradients coincide.
Divergences themselves also vanish as the distributions approach equality,
so their forward computations can suffer cancellation as well.
Applications potentially affected by these errors include knowledge distillation,
reinforcement learning, and proximal optimization, where these
quantities serve as fitting terms and regularizers.
\Cref{sec:applications} expands on these.

\textbf{Contributions of this paper.}
For nearly equal categorical distributions parameterized by logits,
we identify five numerical issues in standard implementations of the
forward and gradient computations of
1) the forward and reverse KL divergences, including Monte Carlo KL
estimators and entropy penalties (\cref{sec:kl}), and
2) the generalized Jensen--Shannon divergence (\cref{sec:jsd}).
To address these issues, we propose eight reformulations and provide
executable JAX examples validated by property-based tests.
\Cref{tab:numerical-issues} summarizes the settings.

\textbf{Does it matter in practice?}
The examples isolate the arithmetic rather than measure large-scale
training outcomes. \Cref{sec:discussion} further discusses their
practical significance.

\begin{table}[t]
\centering\small
\papercaption{Numerical issues for nearly equal categorical distributions parameterized by logits}{}
\label{tab:numerical-issues}
\setlength{\tabcolsep}{3.5pt}
\renewcommand{\arraystretch}{1.12}
\begin{tabular*}{\linewidth}{@{\extracolsep{\fill}}lll@{}}
\toprule
\textbf{Setting} & \textbf{Forward issue} & \textbf{Gradient issue}\\
\midrule
Forward KL (\cref{sec:forward})
 & \hyperref[issue:kl-values]{\textcolor{issueink}{KL-sum cancellation}} & \hyperref[issue:forward]{\textcolor{issueink}{Softmax subtraction}}\\
Reverse KL (\cref{sec:reverse})
 & \hyperref[issue:kl-values]{\textcolor{issueink}{KL-sum cancellation}} & \hyperref[issue:logsoftmax]{\textcolor{issueink}{Log-softmax subtraction}}\\
Entropy \textcolor{black!55}{via reverse KL to uniform} (\cref{sec:entropy})
 & \hyperref[sec:entropy]{\textcolor{black!55}{No issue present}} & \hyperref[sec:entropy]{\textcolor{issueink}{Log-probability centering}}\\
Monte Carlo K1: Log-ratio (\cref{sec:k1})
 & \hyperref[issue:logsoftmax]{\textcolor{issueink}{Log-softmax subtraction}}
 & \hyperref[sec:k1]{\textcolor{black!55}{No issue present}}\\
Monte Carlo K2: Squared log-ratio (\cref{sec:k2})
 & \hyperref[issue:logsoftmax]{\textcolor{issueink}{Log-softmax subtraction}} & \hyperref[sec:k2]{\textcolor{issueink}{Inherits forward issue}}\\
Monte Carlo K3: Control variate (\cref{sec:k3})
 & \hyperref[issue:k3]{\textcolor{issueink}{Remainder cancellation}} & \hyperref[issue:k3]{\textcolor{issueink}{Rounded exponential}}\\
Generalized JSD (\cref{sec:jsd})
 & \begin{tabular}[t]{@{}l@{}}\hyperref[issue:kl-values]{\textcolor{issueink}{KL-sum cancellation}}\\[-2pt]\hyperref[issue:jsd]{\textcolor{issueink}{Mixture log-ratio rounding}}\end{tabular} & \hyperref[issue:jsd]{\textcolor{issueink}{Mixture log-ratio rounding}}\\
\bottomrule
\end{tabular*}
\end{table}

\textbf{Setup and notation.}
We consider a categorical distribution $q$ over $K$ categories parameterized by logits $z\in\R^K$,
so the softmax of the logits gives the distribution, \ie, $q=\softmax(z)$.
We restrict our analysis to this logit parameterization, and the
numerical issues may be different under other parameterizations.
Often in machine learning, the logits are produced by a model $f_\theta$ with
parameters $\theta$, mapping an input $x$ to logits, \ie, $z=f_\theta(x)$.
We also have a fixed reference distribution $p=\softmax(z_{\mathrm{ref}})$
parameterized by logits $z_{\mathrm{ref}}\in\R^K$, potentially also generated by another model.
We analyze the forward computation of each divergence and its
gradient with respect to $z$. Gradients with respect to
$\theta$ follow by the chain rule.
\Cref{app:helpers} gives further details on the numerical conventions,
experimental protocol, and helper implementations.

\newpage
\section{Applications of optimizing divergences}
\label{sec:applications}
Divergences appear in many optimization objectives, both as fitting terms
and as regularizers. We consider three examples where the distributions
can be nearly equal, including at initialization or near an optimum.

\textbf{Knowledge distillation and on-policy distillation.}
Knowledge distillation trains a student to match a teacher's predictions
\citep{bucilua2006compression,hinton2015distilling,kim2016sequence},
with on-policy variants using student-generated contexts
\citep{agarwal2024onpolicy,gu2024minillm}.
The student is trained by solving
\begin{equation}
 \min_\theta\ \E_{x\sim\mu}
 \left[D\bigl(p(\cdot\mid x),q_\theta(\cdot\mid x)\bigr)\right],
 \label{eq:app-opd}
\end{equation}
where $q_\theta(\cdot\mid x)$ and $p(\cdot\mid x)$ are the student and
fixed teacher predictions at context $x$. The context distribution $\mu$
comes from data or student rollouts and is held fixed during each update.
The divergence $D$ is commonly chosen as the forward KL, reverse KL,
or generalized JSD, as in \citet[Section~4.1]{agarwal2024onpolicy}.
Here, these numerical issues can arise near
optimality as the student's predictions approach the teacher's.

\textbf{Reinforcement learning.}
Many RL methods, including InstructGPT \citep{ouyang2022instruct},
optimize the reward along with a KL penalty to a reference policy by solving
\begin{equation}
 \max_\theta\ \E_{y\sim\pi_\theta}[r(y)]
 -\lambda\KL{\pi_\theta}{\pi_{\rm ref}},
 \label{eq:app-rl}
\end{equation}
where the policy $\pi_\theta$ maps contexts to distributions over outputs
$y$, $r(y)$ is the reward, $\pi_{\rm ref}$ is a fixed reference policy,
and $\lambda>0$ is a penalty weight.
The forward KL can also be used between the sampling and training policies.
The numerical issues discussed in this paper could especially arise at
the beginning of training when the policy is initialized close to the
reference policy, although the gradient may be dominated by the main
policy gradient term.

\textbf{Proximal optimization.}
Proximal methods minimize a base objective with a penalty that discourages
large changes from the current iterate. A reverse-KL proximal update,
as in \citet{teboulle1992entropic}, takes the form
\begin{equation}
 \theta_{k+1}\in\arg\min_\theta
 \left\{F(\theta)+\eta_k^{-1}\KL{q_\theta}{q_{\theta_k}}\right\},
 \label{eq:app-proximal}
\end{equation}
where $F$ is the main objective, $\eta_k>0$ is a step-size parameter,
and $q_{\theta_k}$ is held fixed during each update.
Reverse-KL regularization of successive iterates also motivates
natural-gradient methods \citep{amari1998natural,martens2020natural}.
The numerical issues discussed in this paper can arise when successive
distributions are nearly equal, although the gradient may be dominated
by that of the main objective $F$.

\section{Discussion on the scope and practical significance}
\label{sec:discussion}
Before developing the reformulations, we discuss where they apply
and how to interpret their numerical accuracy in optimization.

\textbf{\cautionmark\enspace When to use the reformulations.}
Some reformulations also require nearly equal distributions and may be
numerically unstable outside this regime. Full production implementations
are outside the scope of this paper, but we indicate where guards and
fallback evaluations are needed.
For example, the
\href{https://jonbarron.info/softmaxfix/}{soft cross-entropy implementation}
of \citet{barron2026softmax} uses the correction only for log-ratios
with magnitude below one and switches to direct probability subtraction
otherwise. \Cref{sec:reformulation-pitfalls} illustrates this limitation
for forward~KL: although the reformulation is algebraically valid, its
intermediate computations can overflow away from equality. The same
limitation applies to other reformulations that share this log-ratio
reconstruction.

\textbf{Numerical issues are exacerbated at lower precision.}
Our examples use FP32. \citet[Section~7]{barron2026softmax} compare
BF16, FP32, and FP64 in soft-target distillation and observe larger
residual gradient errors and parameter drift in BF16. Lower precision
can make rounding dominate at larger separations between distributions,
when the divergence and its gradient are larger. Other lower-precision
formats, such as FP8 and FP4, could make these issues even more prevalent
when used to compute divergences and their gradients. At higher
precision, large relative errors may still have very small absolute
magnitude.

\textbf{Adam's stabilizing constant.}
\citet[Sections~5 and~7]{barron2026softmax} study how the stabilizing
constant $\epsilon$ in Adam \citep{kingma2015adam} limits the amplification
of small gradients. Our examples examine numerical accuracy without
measuring its effects on training. In their soft-target distillation
experiment, Adam amplifies rounding residuals and moves a student away
from an initially identical teacher. The divergences considered here
raise a similar concern: their true gradients vanish at equality, but
numerical errors can produce nonzero updates. The practical effect
depends on the parameter-gradient statistics used by Adam, after
backpropagation through the model and combination with other objective
terms.

\textbf{Remaining rounding errors.}
Accuracy also depends on normalization, reductions, and input storage.
A reformulation cannot recover a displacement already rounded away in
its inputs. Compensated summation can retain rounding residuals in the sampled
log-ratio (\cref{fix:sampled-logratio}), improving accuracy when its two
differences nearly cancel \citep[Sections~3--4]{ogita2005accurate}.
This costs additional arithmetic and cannot correct errors already
present in the supplied log-normalizers.

\section{Fixing Kullback--Leibler divergences}
\label{sec:kl}
From the setup, $p=\softmax(z_{\mathrm{ref}})$ is the fixed categorical
reference and $q=\softmax(z)$ is the distribution being optimized.
The forward and reverse KL are defined as:
\par
\begin{minipage}[t]{.49\linewidth}
\begin{definition}[Forward KL]
\label{def:forward}
\begin{equation}
 \begin{aligned}
 \KL{p}{q}&=\E_{a\sim p}[\log p_a-\log q_a]\\
 &\mathrel{\color{black!55}=}{\color{black!55}\sum_i p_i(\log p_i-\log q_i)}.
 \end{aligned}
 \label{eq:forward-loss}
\end{equation}
\end{definition}
\end{minipage}\hfill
\begin{minipage}[t]{.49\linewidth}
\begin{definition}[Reverse KL]
\label{def:reverse}
\begin{equation}
 \begin{aligned}
 \KL{q}{p}&=\E_{a\sim q}[\log q_a-\log p_a]\\
 &\mathrel{\color{black!55}=}{\color{black!55}\sum_i q_i(\log q_i-\log p_i)}.
 \end{aligned}
 \label{eq:reverse-loss}
\end{equation}
\end{definition}
\end{minipage}
\par
The only difference is the order of the arguments. The names
\emph{forward} and \emph{reverse} refer to that order when optimizing
$q$ with $p$ fixed. Both are nonnegative and vanish at $q = p$.
Optimization can favor different solutions when the model family is not
expressive enough, e.g., the forward KL is typically \emph{mode-covering}
and the reverse KL is often \emph{mode-seeking}.

\textbf{Basic derivatives.}
Holding $p$ fixed, differentiating with respect to $z$ gives a probability
difference for the forward KL and a centered log-ratio weighted by $q$
for the reverse KL:
\par
\begin{minipage}[t]{.49\linewidth}
\vspace{0pt}
\begin{tcolorbox}[definitionbox,title={\textbf{Forward KL gradient}}]
\begin{equation}
 \nabla_z\KL{p}{q}=q-p.
 \label{eq:forward-grad}
\end{equation}
\end{tcolorbox}
\end{minipage}\hfill
\begin{minipage}[t]{.49\linewidth}
\vspace{0pt}
\begin{tcolorbox}[definitionbox,title={\textbf{Reverse KL gradient}}]
\begin{equation}
 \nabla_z\KL{q}{p}=q(s-q^\top s),
 \label{eq:logratio-basic}
\end{equation}
{\centering where $s=\log q-\log p$.\par}
\end{tcolorbox}
\end{minipage}
\par

\textbf{This section.}
The two KL directions require different numerical evaluations.
We begin with forward KL and its connection to soft cross-entropy
(\cref{sec:forward}), then turn to reverse KL
(\cref{sec:reverse}). Monte Carlo estimates replace
the KL sums with samples, with expected gradients that depend on the
estimator and sampling direction (\cref{sec:monte-carlo-kl}).
Entropy regularization is another instance of the reverse KL, with a
uniform reference (\cref{sec:entropy}).

\subsection{Fixing the forward KL}
\label{sec:forward}
The forward KL connects directly to the soft cross-entropy analysis of
\citet{barron2026softmax}. With $p$ fixed, the (soft) cross-entropy
$\CE(p,q)=-\sum_i p_i\log q_i$ equals $\KL{p}{q}+H(p)$, where
$H(p)=-\sum_i p_i\log p_i$ is the reference entropy.
Both therefore have gradient $q-p$ (\cref{eq:forward-grad}).
The direct forward computation first subtracts log-probabilities, then
sums their weighted differences. Errors in either step can overwhelm
the small divergence.
We examine these two issues before turning to the shared gradient.
\begin{numericalissue}[Log-softmax subtraction, e.g., as in \cref{eq:forward-loss,eq:reverse-loss}]
\label{issue:logsoftmax}
Evaluating the log-softmax separately for the logits $z$ and
$z_{\mathrm{ref}}$ can round distinct log-probabilities to the same value.
Subtracting the rounded log-softmax outputs can then erase the small entries of
the log-ratio $s=\log q-\log p$. Subtracting the maximum logit
stabilizes normalization but does not prevent this error.
\end{numericalissue}

\citet[Eq.~(9)]{barron2026softmax} compute $s$ from centered logit
differences to avoid this subtraction. For $\delta=z-z_{\mathrm{ref}}$
and $d=\delta-\sum_i p_i\delta_i$, their expression is
\begin{equation}
 s=d-\logp\left(\sum_i p_i\expm(d_i)\right).
 \label{eq:logratio}
\end{equation}
This identity follows from $q/p=e^\delta/\sum_i p_i e^{\delta_i}$.
Subtracting a common scalar from $\delta$ leaves this ratio unchanged.
Choosing $\sum_i p_i\delta_i$ gives $q/p=e^d/\sum_i p_i e^{d_i}$, using the shift
invariance of softmax \citep[Eqs.~(1.3)--(1.4)]{blanchard2021softmax}.
The $\expm$ and $\logp$ routines retain the small increments in
this identity without first rounding $e^d$ or $\sum_i p_i e^{d_i}$ near one.

Even with an accurate $s$, the signed sum
$\KL{p}{q}=-\sum_i p_i s_i$ can still suffer cancellation.
\begin{numericalissue}[KL-sum cancellation, e.g., as in \cref{eq:forward-loss}]
\label{issue:kl-values}
Near equality, each summand satisfies
\begin{equation}
 -p_i s_i=p_i(\log p_i-\log q_i)
 \approx (p_i-q_i)+\frac{(q_i-p_i)^2}{2p_i}.
 \label{eq:kl-summand-expansion}
\end{equation}
The terms $-p_i s_i$ have mixed signs and can be much larger than
their sum: the $p_i-q_i$ contributions cancel because
$\sum_i p_i=\sum_i q_i=1$, leaving the smaller squared terms.
Small errors in the log-ratios, the products, or the
summation need not cancel and can overwhelm the remaining divergence.
\end{numericalissue}

The forward computation therefore needs both an accurate log-ratio
(\cref{issue:logsoftmax}) and a sum that avoids cancellation between
categories (\cref{issue:kl-values}).
Adding $q_i-p_i$ to each summand removes the cancelling contributions
without changing the total, since $\sum_i(q_i-p_i)=0$.
Each corrected summand is $p_i(e^{s_i}-1-s_i)=p_iE_2(s_i)$,
where $E_2(x)=e^x-1-x$.
\begin{reformulation}[Fixes \cref{issue:logsoftmax,issue:kl-values} for the forward KL computation in \cref{eq:forward-loss}]
\label{fix:kl-values}
Use $s$ from \cref{eq:logratio} to avoid subtracting rounded
log-probabilities. Replace the signed KL sum by the nonnegative-remainder
identity:
\begin{equation}
 \KL{p}{q}=\sum_i p_i E_2(s_i).
 \label{eq:kl-values}
\end{equation}
Evaluate $E_2$ near zero using its series expansion to avoid cancellation.
\end{reformulation}
The same correction is used as a control variate in Monte Carlo KL estimation
\citep{schulman2020kl}. Since $e^x\geq1+x$, every remainder
$E_2(s_i)$ is nonnegative, avoiding cancellation between categories.
Computing $E_2(x)$ as $\expm(x)-x$ still subtracts nearly equal
quantities. Near zero, evaluate it directly using a Taylor expansion,
as in the \texttt{expm1x} routine of \citet[pp.~36--37]{maechler2026dpq}:
\begin{equation}
 E_2(x)=x^2\left(\tfrac12+\tfrac x6+\tfrac{x^2}{24}+\cdots\right),
 \qquad E_2'(x)=\expm(x).
 \label{eq:remainder}
\end{equation}
\Cref{app:helpers} gives the \texttt{e2} implementation and its
connection to established remainder evaluations.

\textbf{Gradient computation.}
A similar rounding issue affects the gradient in \cref{eq:forward-grad},
which requires the probability difference $q-p$. Separate softmax
evaluations can round nearby probabilities to the same value.
\begin{numericalpair}
\begin{numericalissue}[Softmax subtraction, e.g., as in \cref{eq:forward-grad}]
\label{issue:forward}
The difference $q-p=\softmax(z)-\softmax(z_{\mathrm{ref}})$ subtracts
softmax outputs, whereas \cref{issue:logsoftmax} subtracts log-softmax
outputs. Either subtraction can erase a small difference when the
outputs are rounded separately, even if each output is individually
accurate. Differentiating the direct KL sum through log-softmax produces
the probability difference $q-p$. Normalization roundoff can leave
a residual at identical logits.
\end{numericalissue}
The retained log-ratio recovers the small probability difference that
softmax subtraction can erase (\cref{issue:forward}).
The final correction uses the gradient's zero-sum identity to reduce
accumulated roundoff.
\begin{reformulation}[Fixes \cref{issue:forward} for the forward KL gradient in \cref{eq:forward-grad}]
\label{fix:forward}
The soft-target correction of \citet{barron2026softmax} evaluates
$q-p$ from the reconstructed $s$ in \cref{eq:logratio} and corrects
its numerical sum residual:
\begin{equation}
 u=p\expm(s),\qquad
 \nabla_z\KL{p}{q}=u-q(\sum_i u_i).
 \label{eq:forward-fix}
\end{equation}
In exact arithmetic, $u=q-p$ and $\sum_i u_i=0$. The final term
reduces the rounding residual.
\end{reformulation}
\end{numericalpair}

\Cref{lst:forward_kl} combines the remainder with this explicit gradient.
\Cref{fig:example-forward_kl} tests both as the logits approach equality.
The helper \texttt{categorical} implements \cref{eq:logratio} and returns
$p,q,\log p,\log q,s$ (\cref{app:helpers}).
These examples focus on numerical accuracy. An optimized implementation
can avoid computing quantities that the loss does not need.
Ordinary autodiff of the complete remainder-based graph is also accurate in
the tested sweep. Retaining the ordinary cross-entropy sum instead needs
the backward correction despite an accurate forward cross-entropy.

\examplepair
  {\begin{inlinepaperlisting}[listingcomparison]{forward_kl}{Forward KL example code}
\begin{minipage}[t]{.485\linewidth}
\codeheading{issueink}{Basic}
\begin{lstlisting}[style=paperpython,firstnumber=1]
def forward_kl_basic(z, z_ref):
    p, _, lp, lq, _ = categorical(z, z_ref)
    return jnp.sum(p * (lp - lq))
\end{lstlisting}
\end{minipage}\hfill
\begin{minipage}[t]{.485\linewidth}
\codeheading{clearink}{Fixed}
\begin{lstlisting}[style=paperpython,firstnumber=1]
@custom_gradient
def forward_kl_fixed(z, z_ref):
    assert_close(z, z_ref)
    p, q, _, _, s = categorical(z, z_ref)
    u = p * jnp.expm1(s)
    value = jnp.sum(p * e2(s))
    gradient = u - q * jnp.sum(u)
    return value, gradient
\end{lstlisting}
\end{minipage}
\end{inlinepaperlisting}
}
  {\centering
\includegraphics[width=1.0\linewidth,trim=0 4pt 0 0,clip]{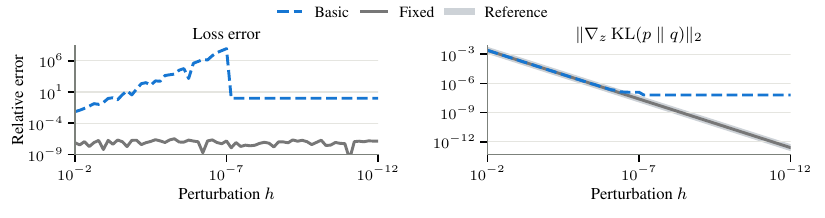}
\papercaption[figure]{Rounding and cancellation distort forward KL and its gradient.}{This example uses fixed reference logits $z_{\mathrm{ref}}=(-1,0,1)$ and varying logits $z=(-1,h,1)$. The reformulation retains the divergence and gradient as $q$ approaches $p$.}
\label{fig:example-forward_kl}
}

\subsubsection{Pitfalls of the reformulations}
\label{sec:reformulation-pitfalls}
The full-distribution KL and JSD reformulations share the log-ratio
reconstruction in \cref{eq:logratio}. Its exponentials can overflow away
from equality, and its logit subtraction can erase small differences even
near equality. We illustrate both limitations with forward KL. The
closeness guard addresses the first but does not resolve the second.

\textbf{\cautionmark\enspace Overflow away from equality.}
The log-ratio identity in \cref{eq:logratio} holds for finite logits,
but its exponential evaluation requires a guard restricting its use
to nearly equal distributions. Otherwise, intermediate quantities can
overflow even when the desired result is finite.
With $z_{\mathrm{ref}}=(-h,0)$ and $z=(h,0)$, the true
log-ratio is $s=(h,-h)$. At $h=50$, the reconstruction forms an
exponential near $e^{100}$ and overflows in FP32, although the log-ratio
and KL are finite (\cref{fig:range-limits}). A small KL does not bound every probability ratio.

The implementations use \texttt{assert\_close} (\cref{app:helpers})
to restrict this evaluation to nearly equal distributions.
Full-range use requires a fallback, such as the hybrid evaluation of
\citet{barron2026softmax}.

\begin{figure}[t]
\input{examples/results/range_limits}
\par\vspace{8pt}
\captionsetup{type=figure}
\input{examples/results/logit_rounding}
\end{figure}

\begin{samepage}
\textbf{\cautionmark\enspace Rounding can remain near equality.}
Take $z_{\mathrm{ref}}=(0,h)$ and $z=(1,1)$ with $h=2^{-25}$.
The inputs retain $h$ exactly, but FP32 subtraction gives
$\delta=z-z_{\mathrm{ref}}=(1,1)$. Centering this constant vector gives
zero, so the fixed forward KL and its gradient both vanish even
though $p\ne q$ (\cref{fig:example-logit_rounding}). The reverse-KL and
JSD evaluations share this failure because they use the same logit differences.
It occurs before the stable elementary functions are applied and passes
the \texttt{assert\_close} check.
The reconstruction avoids subtracting separately rounded log-probabilities,
but its own logit subtraction can still erase the signal.
Max and mean centering, anchoring to one category, and aligning the logit
offsets caused regressions or left rounding failures unresolved in our tests.
Finding a simple FP32
reformulation without these trade-offs remains an open problem.
\par
\end{samepage}

\subsection{Fixing the reverse KL}
\label{sec:reverse}
The reverse KL averages under $q$, which depends on the logits $z$.
Its forward computation inherits log-softmax subtraction
(\cref{issue:logsoftmax}) and KL-sum cancellation (\cref{issue:kl-values}).
With $p$ and $q$ exchanged in \cref{eq:kl-summand-expansion}, the signed
differences $q_i-p_i$ again sum to zero. The remainder identity in
\cref{eq:kl-values} applies with $-s$ in place of $s$.
\begin{reformulation}[Fixes \cref{issue:logsoftmax,issue:kl-values} for the reverse KL computation in \cref{eq:reverse-loss}]
\label{fix:reverse-value}
Adding $q_i(e^{-s_i}-1)=p_i-q_i$ to each summand $q_i s_i$
leaves the total unchanged because $\sum_i(p_i-q_i)=0$.
Using $s$ from \cref{eq:logratio}, this replaces the signed sum in
\cref{eq:reverse-loss} with nonnegative remainders:
\begin{equation}
 \KL{q}{p}=\sum_i q_i E_2(-s_i).
\label{eq:reverse-value}
\end{equation}
Evaluate $E_2$ with \cref{eq:remainder} near zero.
\end{reformulation}

\examplepair
  {\begin{inlinepaperlisting}[listingcomparison]{reverse_kl}{Reverse KL example code}
\begin{minipage}[t]{.485\linewidth}
\codeheading{issueink}{Basic}
\begin{lstlisting}[style=paperpython,firstnumber=1]
def reverse_kl_basic(z, z_ref):
    _, q, lp, lq, _ = categorical(z, z_ref)
    return jnp.sum(q * (lq - lp))
\end{lstlisting}
\end{minipage}\hfill
\begin{minipage}[t]{.485\linewidth}
\codeheading{clearink}{Fixed}
\begin{lstlisting}[style=paperpython,firstnumber=1]
@custom_gradient
def reverse_kl_fixed(z, z_ref):
    assert_close(z, z_ref)
    _, q, _, _, s = categorical(z, z_ref)
    delta = z - z_ref
    delta = delta - delta[0]
    value = jnp.sum(q * e2(-s))
    centered = delta - jnp.sum(q * delta)
    gradient = q * centered
    gradient = gradient - q * jnp.sum(gradient)
    return value, gradient
\end{lstlisting}
\end{minipage}
\end{inlinepaperlisting}
}
  {\centering
\includegraphics[width=1.0\linewidth,trim=0 4pt 0 0,clip]{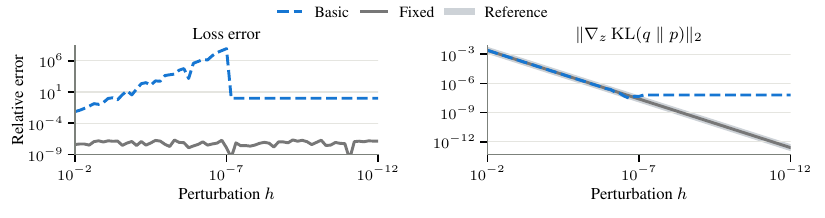}
\papercaption[figure]{Centering logit differences preserves the reverse-KL gradient.}{This example uses fixed reference logits $z_{\mathrm{ref}}=(-1,0,1)$ and varying logits $z=(-1,h,1)$. The reformulation retains the divergence and gradient as $q$ approaches $p$.}
\label{fig:example-reverse_kl}
}

\textbf{Gradient computation.}
The reverse-KL gradient in \cref{eq:logratio-basic} centers the log-ratio
$s$, whose small entries can be erased by log-softmax subtraction
(\cref{issue:logsoftmax}).
The log-normalizer offset is common to every category and cancels under
centering. Removing it before evaluation avoids rounding away the logit
differences that remain.
Differentiating the exponential-family KL identity using the log-partition
derivatives of
\citet[Eq.~(5.10) and Proposition~3.1]{wainwright2008exponential}
gives the centered gradient in terms of the logit differences
$\delta=z-z_{\mathrm{ref}}$.
\begin{reformulation}[Fixes \cref{issue:logsoftmax} for the reverse KL gradient in \cref{eq:logratio-basic}]
\label{fix:reverse}
Center the logit differences instead of the separately computed log-ratio:
\begin{equation}
 u=q\Bigl(\delta-\sum_i q_i\delta_i\Bigr),\qquad
 \nabla_z\KL{q}{p}=u-q(\sum_i u_i).
 \label{eq:reverse-fix}
\end{equation}
In the implementation, first subtract $\delta_1$ from every entry of
$\delta$, as in \cref{lst:reverse_kl}. This removes a common logit offset
before centering. The final term applies the zero-sum correction from
\cref{eq:forward-fix}.
\end{reformulation}
The following proposition shows why centering the logit differences in
\cref{eq:reverse-fix} gives the same gradient as centering the log-ratio in
\cref{eq:logratio-basic}.
\begin{proposition}[Centering log-ratios and logit differences]
\label{prop:reverse-centering}
For finite logits $z,z_{\mathrm{ref}}\in\R^K$, let $q=\softmax(z)$ and
$p=\softmax(z_{\mathrm{ref}})$. Define the log-ratio $s=\log q-\log p$
and the logit differences $\delta=z-z_{\mathrm{ref}}$. Then
\begin{equation}
 s-\sum_i q_i s_i
 =\delta-\sum_i q_i\delta_i.
 \label{eq:reverse-centering}
\end{equation}
\end{proposition}
\begin{samepage}
\begin{proof}
Expanding $s_i=\log q_i-\log p_i$ using the two softmax distributions gives
\[
 s_i=\delta_i+\log\sum_j e^{z_{\mathrm{ref},j}}-\log\sum_j e^{z_j}.
\]
The log-normalizer difference is the same for every category. Since
$\sum_jq_j=1$, subtracting the $q$-weighted mean removes this common offset:
\[
 s_i-\sum_jq_js_j=\delta_i-\sum_jq_j\delta_j.\qedhere
\]
\end{proof}
\end{samepage}

Subtracting a common scalar from every logit difference also leaves the
centered gradient unchanged in exact arithmetic. \Cref{lst:reverse_kl}
uses $\delta_1$ for convenience, avoiding an additional maximum reduction.
Subtracting $\max_i\delta_i$ is also valid. Either choice bounds the shifted
magnitudes by the span of $\delta$, and this centering step does not require
nonpositive entries.

In exact arithmetic, the centered vector $u$ satisfies $\sum_i u_i=0$,
so the final correction in \cref{eq:reverse-fix} leaves the gradient
unchanged. The reverse-KL gradient is also zero whenever $q=p$,
including when the logits differ by a common offset. In floating point,
the choice of offset and the accuracy of normalization and centering
still affect the result. Computed probabilities need not sum to exactly
one, so a common shift can leave a residual proportional to the
normalization error. The final correction reduces this residual with
an extra sum.

\Cref{lst:reverse_kl} reconstructs $s$ for the forward remainder and applies
the two centering steps for the gradient. \Cref{fig:example-reverse_kl} compares these
evaluations with the direct KL sum. Ordinary autodiff of the complete remainder-based
forward graph also passes the tests. The explicit rule gives a shorter
derivative and supports other formulas.

\subsubsection{Fixing the entropy gradient}
\label{sec:entropy}
Entropy regularization is a special case of reverse KL. For $K$ categories
and a uniform reference $u_i=1/K$, entropy satisfies the identity
\citep{cover2006information}:
\begingroup
\setlength{\belowdisplayskip}{3pt plus 1pt minus 1pt}
\setlength{\belowdisplayshortskip}{3pt plus 1pt minus 1pt}
\begin{equation}
 H(q)=\log K-\KL{q}{u}
 =-\sum_iq_i\log q_i.
 \label{eq:entropy-uniform}
\end{equation}
\endgroup
Differentiating this identity gives
$\nabla_z H(q)=-\nabla_z\KL{q}{u}$, so \cref{fix:reverse} applies
with a uniform reference and a sign change.
Near uniformity, the entropy stays near $\log K$ while its gradient
vanishes. Centering separately computed log-probabilities can round
this small gradient to zero, just as for reverse KL. For $z=(0,10^{-8})$,
directly evaluating and differentiating the entropy sum in FP32
returns an accurate entropy near $\log2$ but rounds its nonzero
gradient to zero.
\Cref{lst:categorical_entropy} specializes the reverse-KL gradient rule,
including its final zero-sum correction (\cref{eq:reverse-fix}), while
keeping the usual entropy sum for the forward computation. \Cref{fig:example-categorical_entropy}
illustrates the resulting gradient accuracy near uniformity.
The entropy deficit $\log K-H(q)$, if needed, should be evaluated
through the reverse-KL remainder in \cref{eq:reverse-value}.

\textbf{A practical note.}
This issue may be less consequential in practice when entropy is used as
a regularizer. The numerical failure studied here occurs near the uniform
distribution, whereas the main task may favor nonuniform predictions.
Accurately retaining this small entropy gradient may therefore matter
less than retaining a divergence gradient near a teacher or reference
distribution.

\examplepair
  {\begin{inlinepaperlisting}[listingcomparison]{categorical_entropy}{Entropy example code}
\begin{minipage}[t]{.485\linewidth}
\codeheading{issueink}{Basic}
\begin{lstlisting}[style=paperpython,firstnumber=1]
def categorical_entropy_basic(z):
    log_q = jax.nn.log_softmax(z)
    q = jnp.exp(log_q)
    return -jnp.sum(q * log_q)
\end{lstlisting}
\end{minipage}\hfill
\begin{minipage}[t]{.485\linewidth}
\codeheading{clearink}{Fixed}
\begin{lstlisting}[style=paperpython,firstnumber=1]
@custom_gradient
def categorical_entropy_fixed(z):
    log_q = jax.nn.log_softmax(z)
    q = jnp.exp(log_q)
    value = -jnp.sum(q * log_q)
    a = z - z[0]
    centered = a - jnp.sum(q * a)
    gradient = -q * centered
    gradient = gradient - q * jnp.sum(gradient)
    return value, gradient
\end{lstlisting}
\end{minipage}
\end{inlinepaperlisting}
}
  {\centering
\includegraphics[width=0.5\linewidth,trim=0 4pt 0 0,clip]{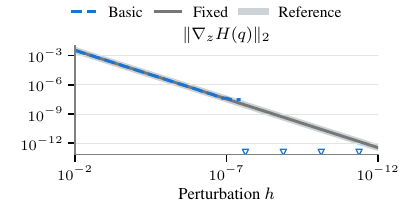}
\papercaption[figure]{An accurate entropy value can accompany an inaccurate gradient.}{This example uses two-category logits $z=(0,h)$, so $q$ approaches the uniform distribution as $h$ approaches zero. Reversing the fixed reverse-KL gradient retains the small derivative that ordinary autodiff rounds to zero.}
\label{fig:example-categorical_entropy}
}

\subsection{Fixing Monte Carlo KL estimators}
\label{sec:monte-carlo-kl}
Summing over every category can be expensive when the distribution is
large. Monte Carlo estimation instead evaluates log-probabilities at
sampled categories, a longstanding approach to KL estimation
\citep{carlin1992montecarlo,hershey2007approximating}.
At a fixed context, write $s_a=\log q_a-\log p_a$ for the log-ratio
at a sampled category $a$.
Three commonly used estimators, as notated by \citet{schulman2020kl}, are:

\begingroup
\tcbset{groupeddefinition}
\begin{definition}[Monte Carlo KL estimators]
\label{def:sampled-estimators}
\begin{tcbraster}[estimatorrow]
\begin{tcolorbox}[title={K1: Log-ratio}]
$\mathstrut k_1(s_a)=s_a$
\end{tcolorbox}
\begin{tcolorbox}[title={K2: Squared log-ratio}]
$\mathstrut k_2(s_a)=s_a^2/2$
\end{tcolorbox}
\begin{tcolorbox}[title={K3: Control variate}]
$\mathstrut k_3(s_a)=s_a+e^{-s_a}-1$
\end{tcolorbox}
\end{tcbraster}
\end{definition}
\endgroup

For a minibatch of $N$ independent categories $a_n\sim q$, the Monte Carlo
estimate to the reverse KL is
\begin{equation}
 \frac{1}{N}\sum_{n=1}^{N}k(s_{a_n}),
 \label{eq:sampled-minibatch}
\end{equation}
where $k$ is any of the three estimators. The Monte Carlo estimate to the
forward KL instead uses samples $a_n\sim p$ and evaluates $k(-s_{a_n})$.
When this average is used
as an optimization objective, autodiff differentiates it while treating
the sampled categories as fixed. The choice of estimator and sampling
direction determines whether the value and gradient estimates are
unbiased, as discussed in \cref{sec:sampled-directions}.

\subsubsection{Forward and reverse Monte Carlo KL estimators}
\label{sec:sampled-directions}
\textbf{Expected values.}
The Monte Carlo estimates to the reverse and forward KL have expected values
$\E_{a\sim q}[k(s_a)]$ and $\E_{a\sim p}[k(-s_a)]$, respectively.
K1 and K3 are unbiased in both directions, whereas K2 matches the
corresponding KL only to second order near equality \citep{schulman2020kl}.
In on-policy distillation, sampling from $q$ is called
\emph{student-sampled} estimation because $q$ is the student distribution
\citep{lu2025onpolicy}.

\textbf{Expected gradients.}
\Cref{tab:sampled-gradients} compares the expected gradients in exact arithmetic.
The reverse KL depends on $z$ through both the log-ratio and the sampling
distribution $q$. Treating the sampled categories as fixed omits the
score-function gradient associated with changes in the sampling probabilities.
Thus, differentiating an unbiased forward estimate may not give an
unbiased gradient estimate. With forward-KL sampling, $p$ is fixed, so
there is no sampling-distribution derivative to omit.
\begin{table}[t]
\centering\small
\begin{threeparttable}
\papercaption{Expected gradients of Monte Carlo KL estimators}{}
\label{tab:sampled-gradients}
\renewcommand{\arraystretch}{1.4}
\setlength{\tabcolsep}{17pt}
\begin{tabular}{@{}ccc@{}}
\toprule
\textbf{Estimator}
 & \textbf{Reverse KL} \textcolor{black!55}{$\bigl(\E_{a\sim q}[\nabla_z k_j(s_a)]\bigr)$}
 & \textbf{Forward KL} \textcolor{black!55}{$\bigl(\E_{a\sim p}[\nabla_z k_j(-s_a)]\bigr)$}\\
\midrule
$k_1$ & $0$\enspace\cautionmark[issueink] & $\nabla_z\KL{p}{q}$\\
$k_2$ & $\nabla_z\KL{q}{p}$ & $p s-q\sum_a p_a s_a$\\
$k_3$ & $\nabla_z\KL{p}{q}$ & $\nabla_z\KL{p}{q}$\\
\bottomrule
\end{tabular}
\begin{tablenotes}[flushleft]
\color{black!55}
\item[] \cautionmark[issueink]\enspace K1 should not be directly differentiated as a
reverse-KL penalty: its expected gradient is zero.
\end{tablenotes}
\end{threeparttable}
\end{table}

The reverse-KL sampling identities are established by
\citet[Section~3]{tang2025pitfalls}. For forward-KL sampling, differentiating
$\tfrac12\sum_a p_a s_a^2$ gives the K2 entry, which generally differs
from either KL gradient. K2 therefore minimizes squared log-probability
errors under the reference distribution, with a gradient that matches the
forward-KL gradient to first order near equality.
K1 and K3 give unbiased forward-KL gradients.

\textbf{Applications in reinforcement learning for language models.}
In reinforcement learning for language models, large vocabularies
motivate using these estimators. For example, Kimi k1.5 uses K2 in a
surrogate update for a reverse-KL-regularized objective
\citep[Section~2.3.2, Eqs.~(2)--(3)]{kimi2025k15}, while DeepSeekMath uses
K3 for reverse-KL regularization in GRPO \citep[Eq.~(4)]{shao2024deepseekmath}.
StableReinforce uses forward KL from the sampler to the learner,
with K1 by default or K3
\citep[Sampler--learner KL]{thinkingmachines2026stablereinforce}.

\examplepair
  {\begin{inlinepaperlisting}[listingcomparison]{sampled_log_ratio}{Sampled log-ratio example code}
\begin{minipage}[t]{.485\linewidth}
\codeheading{issueink}{Basic}
\begin{lstlisting}[style=paperpython,firstnumber=1]
def sampled_log_ratio_basic(
    z_a, log_Z, z_ref_a, log_Z_ref
):
    log_q_a = z_a - log_Z
    log_p_a = z_ref_a - log_Z_ref
    return log_q_a - log_p_a
\end{lstlisting}
\end{minipage}\hfill
\begin{minipage}[t]{.485\linewidth}
\codeheading{clearink}{Fixed}
\begin{lstlisting}[style=paperpython,firstnumber=1]
def sampled_log_ratio_fixed(
    z_a, log_Z, z_ref_a, log_Z_ref
):
    z_diff = z_a - z_ref_a
    log_Z_diff = log_Z - log_Z_ref
    return z_diff - log_Z_diff
\end{lstlisting}
\end{minipage}
\end{inlinepaperlisting}
}
  {\centering
\includegraphics[width=1.0\linewidth,trim=0 4pt 0 0,clip]{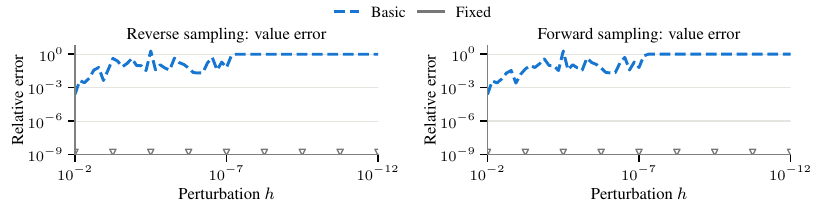}
\papercaption[figure]{Regrouping reduces K1's log-ratio error in both directions.}{Left: reverse-KL sampling. Right: forward-KL sampling. Exact averages for $z_{\mathrm{ref}}=(-1,0,1)$ and $z=(-1,h,1)$. Value errors isolate arithmetic at the supplied FP32 normalizers. }
\label{fig:example-k1_expected}
}

\subsubsection{Shared log-ratio inputs}
\label{sec:sampled-logratio}
Both sampling directions use the same scalar log-ratio and remainder
operations. We evaluate the numerical issues in both directions and compare
the results with each estimator's expected value and gradient to separate
numerical error from estimator bias.
The listings show Monte Carlo estimates to the reverse KL. Forward-KL
implementations follow by negating the log-ratio before evaluating and
differentiating the estimate. K2 uses the same code in either direction.

The log-ratio uses the student logit $z_a$ and reference logit
$z_{\mathrm{ref},a}$ for the same sampled category $a$, together with
their log-normalizers $\log Z=\log\sum_i e^{z_i}$ and
$\log Z_{\mathrm{ref}}=\log\sum_i e^{z_{\mathrm{ref},i}}$ over all
$K$ categories.
Both use a numerically stable log-sum-exp evaluation that shifts by the
maximum logit to avoid overflow \citep{blanchard2021softmax}.
These four scalars determine the sampled log-ratio:
\begin{equation}
 s_a=(z_a-z_{\mathrm{ref},a})-(\log Z-\log Z_{\mathrm{ref}}).
 \label{eq:sampled-logratio}
\end{equation}
Each model computes its own log-normalizer, so evaluating the penalty
requires only the sampled logit and log-normalizer from each model.
This also reduces communication when the models run
separately, such as in distributed setups. Each supplies two scalars per
sampled category rather than its full $K$-dimensional logit vector. Both student inputs
remain differentiable. Throughout, the \emph{input derivatives} of an
estimator are its partial derivatives with respect to these two student
inputs, $z_a$ and $\log Z$, holding the sampled category and the
reference inputs fixed.

All three estimators inherit the log-softmax subtraction issue
(\cref{issue:logsoftmax}): separately computing $z_a-\log Z$ and
$z_{\mathrm{ref},a}-\log Z_{\mathrm{ref}}$ can round the two
log-probabilities to the same value before subtraction. Regrouping the
four scalar inputs avoids forming these intermediate log-probabilities.
\begin{reformulation}[Fixes \cref{issue:logsoftmax} for the sampled log-ratio in \cref{eq:sampled-logratio}]
\label{fix:sampled-logratio}
Evaluate \cref{eq:sampled-logratio} by first computing
$z_a-z_{\mathrm{ref},a}$ and $\log Z-\log Z_{\mathrm{ref}}$, then
subtracting these differences without forming the log-probabilities.
Like the full-distribution reconstruction in \cref{eq:logratio}, this
forms logit differences before log-probabilities. It retains small logit
changes that separate log-probability evaluations can round away.
\end{reformulation}
Stable log-sum-exp still rounds the supplied log-normalizers, and
the two differences can introduce further rounding.
The regrouping therefore does not guarantee small relative error for
every nonzero log-ratio.

For $z_a=h$, $z_{\mathrm{ref},a}=0$, and
$\log Z=\log Z_{\mathrm{ref}}=1$, the log-ratio is $s_a=h$.
At $h=10^{-8}$, the two log-probabilities round to the same value,
so direct subtraction returns zero, whereas \cref{lst:sampled_log_ratio}
retains $h$.

\begin{samepage}
\textbf{Numerical evaluation.}
The three estimators use the same example, with
$z_{\mathrm{ref}}=(-1,0,1)$ and $z=(-1,h,1)$.
All four inputs and both log-normalizer computations use FP32.
To isolate numerical error from sampling noise, we compute the expected
value and gradient by summing over every category. For reverse-KL sampling,
this gives
\begin{equation}
 \E_{a\sim q}[k(s_a)]=\sum_a q_a k(s_a),\qquad
 \E_{a\sim q}[\nabla_z k(s_a)]=\sum_a q_a\,\nabla_z k(s_a).
 \label{eq:enumerated-gradient}
\end{equation}
\end{samepage}
For forward-KL sampling, the weights are $p_a$ and the estimates are
$k(-s_a)$. The value panels compare arithmetic at the supplied FP32
normalizers. The gradient panels compare against the exact distributions
defined by the stored logits, including errors from the normalizers and
their derivatives.

\examplepair
  {\begin{inlinepaperlisting}[listingcomparison]{k2}{Squared log-ratio (K2) example code}
\begin{minipage}[t]{.485\linewidth}
\codeheading{issueink}{Basic}
\begin{lstlisting}[style=paperpython,firstnumber=1]
def reverse_k2_basic(
    z_a, log_Z, z_ref_a, log_Z_ref
):
    s = sampled_log_ratio_basic(
        z_a, log_Z, z_ref_a, log_Z_ref)
    return s * s / 2
\end{lstlisting}
\end{minipage}\hfill
\begin{minipage}[t]{.485\linewidth}
\codeheading{clearink}{Fixed}
\begin{lstlisting}[style=paperpython,firstnumber=1]
def reverse_k2_fixed(
    z_a, log_Z, z_ref_a, log_Z_ref
):
    s = sampled_log_ratio_fixed(
        z_a, log_Z, z_ref_a, log_Z_ref)
    return s * s / 2
\end{lstlisting}
\end{minipage}
\end{inlinepaperlisting}
}
  {\centering
\includegraphics[width=1.0\linewidth,trim=0 4pt 0 0,clip]{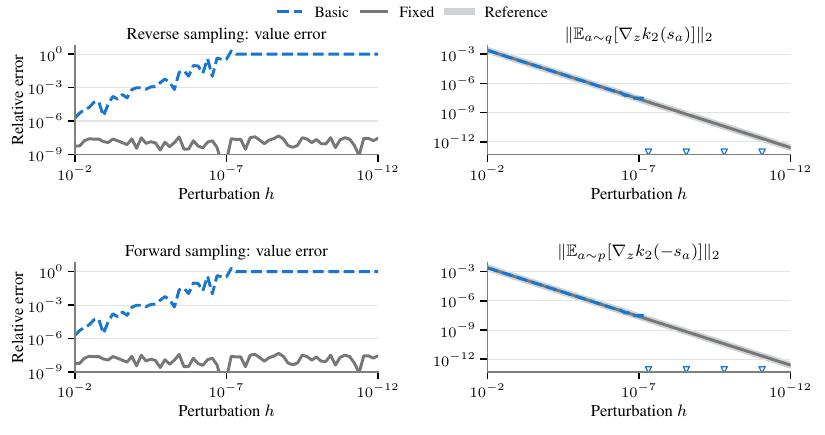}
\papercaption[figure]{K2 inherits log-ratio error in both sampling directions.}{First row: reverse-KL sampling. Second row: forward-KL sampling. Exact averages for $z_{\mathrm{ref}}=(-1,0,1)$ and $z=(-1,h,1)$. Value errors isolate arithmetic at the supplied FP32 normalizers. Gradient references are the expected derivatives of each estimator under the exact distributions.}
\label{fig:example-k2_expected}
}

\subsubsection{Fixing the sampled log-ratio estimator (K1)}
\label{sec:k1}
K1 returns the sampled log-ratio directly, inheriting the rounding issue
in \cref{issue:logsoftmax}. The Monte Carlo estimate to the forward KL
negates this log-ratio.
The input derivatives are constant in both directions, so the regrouping
introduces no change in the gradient computation. Under sampling from $q$,
K1's expected logit gradient remains zero. Under sampling from $p$, it is
the forward-KL gradient in exact arithmetic (\cref{tab:sampled-gradients}).
FP32 backpropagation through the student normalizer can still make this
small expected gradient inaccurate. The log-ratio regrouping does not
correct that separate error.

\Cref{fig:example-k1_expected} shows the reduced log-ratio error, but
regrouping cannot recover differences already rounded away in the supplied
normalizers. For sufficiently small $h$ in the figure's example, both
normalizers round to the same value, leaving a log-ratio of $h$ only for
the perturbed category. In both directions, the expected estimate then scales linearly with
$h$ in magnitude, whereas the true KL is quadratic. The figure's reference holds
the supplied normalizers fixed, isolating error in the subsequent
log-ratio calculation.

\examplepair
  {\begin{inlinepaperlisting}[listingcomparison]{k3}{Control-variate estimator (K3) example code}
\begin{minipage}[t]{.485\linewidth}
\codeheading{issueink}{Basic}
\begin{lstlisting}[style=paperpython,firstnumber=1]
def reverse_k3_basic(
    z_a, log_Z, z_ref_a, log_Z_ref
):
    s = sampled_log_ratio_basic(
        z_a, log_Z, z_ref_a, log_Z_ref)
    return jnp.expm1(-s) + s
\end{lstlisting}
\end{minipage}\hfill
\begin{minipage}[t]{.485\linewidth}
\codeheading{clearink}{Fixed}
\begin{lstlisting}[style=paperpython,firstnumber=1]
@custom_gradient(argnums=(0, 1))
def reverse_k3_fixed(
    z_a, log_Z, z_ref_a, log_Z_ref
):
    s = sampled_log_ratio_fixed(
        z_a, log_Z, z_ref_a, log_Z_ref)
    assert_small_log_ratio(s)
    g = -jnp.expm1(-s)
    return e2(-s), (g, -g)
\end{lstlisting}
\end{minipage}
\end{inlinepaperlisting}
}
  {\centering
\includegraphics[width=1.0\linewidth,trim=0 4pt 0 0,clip]{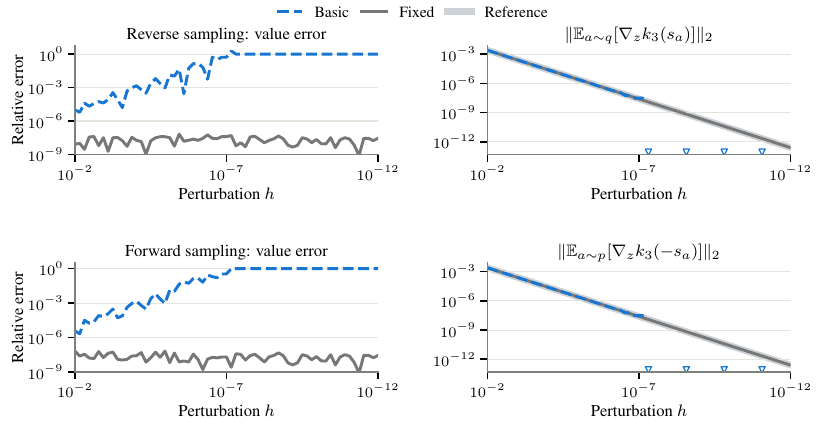}
\papercaption[figure]{K3 needs a stable remainder and derivative in both directions.}{First row: reverse-KL sampling. Second row: forward-KL sampling. Exact averages for $z_{\mathrm{ref}}=(-1,0,1)$ and $z=(-1,h,1)$. Value errors isolate arithmetic at the supplied FP32 normalizers. Gradient references are the expected derivatives of each estimator under the exact distributions.}
\label{fig:example-k3_expected}
}

\subsubsection{Fixing the squared log-ratio estimator (K2)}
\label{sec:k2}
K2 in \cref{def:sampled-estimators} squares the shared log-ratio from
\cref{eq:sampled-logratio}. Rounding in this input propagates to both the penalty
and its two input derivatives.

Rounding the log-ratio to zero in the scalar example of
\cref{sec:sampled-logratio} also erases K2 and both input derivatives.
\Cref{lst:k2} preserves them by regrouping the
log-ratio before squaring it. The square is unchanged by reversing the
log-ratio, but the sampling weights differ.
Given an accurate log-ratio,
neither the square nor its derivative introduces further cancellation,
although sufficiently small outputs can underflow.
\Cref{fig:example-k2_expected} compares both sampling directions with
their respective expected gradients in \cref{tab:sampled-gradients}.
For sampling from $q$, if the score gradients $\nabla_z\log q_a$ are evaluated exactly, a
log-normalizer error shared by every category cancels from this expected
gradient because $\E_{a\sim q}[\nabla_z\log q_a]=0$. It still changes
the expected penalty.

\subsubsection{Fixing the control-variate estimator (K3)}
\label{sec:k3}
The K3 estimator in \cref{def:sampled-estimators} also appears as an
estimator of the extended KL divergence in
\citet[Section~4.1, Eq.~(102)]{nielsen2018montecarlo}, with $p$ and $q$
interchanged. The extended KL includes the terms $p_i-q_i$, which sum
to zero for normalized distributions.
\citet{schulman2020kl} explains the same correction as a zero-mean
control variate added to K1.

An accurate log-ratio alone does not prevent cancellation in K3.
For reverse-KL sampling, its basic evaluation and derivatives are
\begin{equation}
 k_3(s_a)=\expm(-s_a)+s_a,\qquad
 \frac{\partial k_3}{\partial z_a}=1-e^{-s_a},\qquad
 \frac{\partial k_3}{\partial\log Z}=e^{-s_a}-1.
 \label{eq:k3-basic}
\end{equation}
Although $\expm$ retains the small exponential increment, adding $s_a$
still subtracts nearly equal quantities:
$\expm(-s_a)=-s_a+s_a^2/2+O(s_a^3)$, so the linear terms cancel,
leaving a much smaller quadratic remainder.
For forward-KL sampling, the same cancellation occurs in
$k_3(-s_a)=\expm(s_a)-s_a$. Its input derivatives are $e^{s_a}-1$ and
$1-e^{s_a}$.

\textbf{Gradient computation.}
The input derivatives in \cref{eq:k3-basic} also
subtract nearly equal quantities when $e^{-s_a}$ is close to one.
This is the cancellation in $e^x-1$ analyzed by \citet[Section~3]{beebe2002expm1}.

\begin{numericalpair}
\begin{numericalissue}[Remainder cancellation and rounded exponential, e.g., as in \cref{eq:k3-basic}]
\label{issue:k3}
Rounding in $\expm(-s_a)$ can overwhelm the quadratic remainder after
adding $s_a$. In the derivatives, $e^{-s_a}$ can round to one, making
both derivatives zero. Both failures can occur even with an accurate $s_a$.
\end{numericalissue}
To address remainder cancellation and rounding of the exponential in
the derivative (\cref{issue:k3}), evaluate the remainder using
\cref{eq:remainder} and the input derivatives using $\expm$.
\begin{reformulation}[Fixes \cref{issue:k3} for the K3 computation and gradient in \cref{eq:k3-basic}]
\label{fix:k3}
Evaluate the remainder with
the Taylor strategy of \citet[pp.~36--37]{maechler2026dpq} and its
derivative with $\expm$ \citep[Sections~1 and~3]{beebe2002expm1}:
\begin{equation}
 k_3(s_a)=E_2(-s_a),\qquad
 \frac{\partial k_3}{\partial z_a}=-\expm(-s_a),\qquad
 \frac{\partial k_3}{\partial\log Z}=\expm(-s_a).
 \label{eq:k3-gradient}
\end{equation}
For forward-KL sampling, evaluate $k_3(-s_a)=E_2(s_a)$ with input
derivatives $\expm(s_a)$ and $-\expm(s_a)$.
\end{reformulation}
\end{numericalpair}
\Cref{lst:k3} implements the basic and fixed evaluations in
\cref{eq:k3-basic,eq:k3-gradient}. Both take the sampled logits and
log-normalizers, and the fixed version checks $|s_a|\leq 1/2$.

To isolate K3's additional cancellation, take $z_a=-h$,
$z_{\mathrm{ref},a}=-2h$, and $\log Z=\log Z_{\mathrm{ref}}=0$.
The direct log-ratio $s_a=h$ is accurate, but at $h=10^{-8}$ the
direct K3 value and both input derivatives are zero.
The reformulation preserves the small nonzero penalty and both input
derivatives.
Ordinary autodiff of the polynomial remainder also preserves these derivatives.

\textbf{Expected K3 gradients.}
\Cref{fig:example-k3_expected} compares both sampling directions.
Both expected gradients have the forward-KL reference in
\cref{tab:sampled-gradients}, although their expected values estimate
different KL directions.
The fixed update stays close to the reference as the distributions
approach equality, including where the direct update rounds to zero.

\section{Fixing the generalized Jensen--Shannon divergence}
\label{sec:jsd}
The generalized JSD compares $p$ and $q$ with a shared mixture
\citep[Section~IV]{lin1991divergence}, and is used for distillation by
\citet[Section~4.1]{agarwal2024onpolicy}. Its forward
computation inherits the cancellation within each KL term, while its
gradient depends on a small mixture log-ratio. The remainder and
centering identities from the KL sections address these two issues.
\begin{definition}[Generalized JSD]
\label{def:jsd}
For a mixture weight $\beta\in(0,1)$, the generalized JSD is defined by
\begin{equation}
 L_\beta=\beta\KL{p}{m}+(1-\beta)\KL{q}{m}
 \quad\text{where}\quad m=\beta p+(1-\beta)q.
\label{eq:jsd-loss}
\end{equation}
\end{definition}
Dividing by $\beta(1-\beta)$ recovers the forward KL as $\beta\to0$ and
the reverse KL as $\beta\to1$, so the generalized JSD interpolates
between the two KL directions \citep[Section~6, Eqs.~(13)--(16)]{huszar2015train}.

Both KL terms in generalized JSD are susceptible to KL-sum cancellation
(\cref{issue:kl-values}). \citet[Section~II]{raj2008nonnegative} address this issue
using nonnegative series. Here, we apply the KL remainder identity to
each term, which requires accurate log-ratios between the distributions
and their mixture.

Following \citet{barron2026softmax}, we first reconstruct
$s=\log q-\log p$ from logit differences (\cref{eq:logratio}) to avoid
log-softmax subtraction (\cref{issue:logsoftmax}). The remaining quantity
is the mixture log-ratio $h=\log q-\log m$.
Unlike reverse KL, this log-ratio is a nonlinear function of $s$:
$m/q=1+\beta(e^{-s}-1)$. Centering after this transformation cannot
generally remove the common log-normalizer offset in $s$, so replacing
$s$ by $\delta$ changes the gradient. Reconstructing $s$ accurately is
necessary, but the conversion to $h$ can introduce further rounding.
\begin{numericalissue}[Mixture log-ratio rounding, e.g., as in \cref{eq:jsd-loss}]
\label{issue:jsd}
Even with an accurate $s$, forming $m/q=1+\beta(e^{-s}-1)$ can round
the ratio to one, making $h=-\log(m/q)$ zero. Evaluating $\log q-\log m$
instead can subtract separately rounded log-probabilities, as in
\cref{issue:logsoftmax}. Computing $\log m$ with log-sum-exp does not
prevent that subtraction error.
\end{numericalissue}
Use $\expm$ to retain
$e^{-s}-1$ and $\logp$ to take the logarithm without first rounding the
sum with one:
\begin{equation}
 h=-\logp\bigl(\beta\expm(-s)\bigr).
 \label{eq:jsd-fix}
\end{equation}
The two KL terms use $\log(m/p)=s-h$ and $\log(m/q)=-h$.
With these log-ratios retained, cancellation within each signed KL sum
(\cref{issue:kl-values}) remains. The remainder identity avoids this
cancellation.
\begin{reformulation}[Fixes \cref{issue:logsoftmax,issue:kl-values,issue:jsd} for the JSD computation in \cref{eq:jsd-loss}]
\label{fix:jsd-value}
Using the reconstructed $s$ and $h$ from \cref{eq:logratio,eq:jsd-fix},
apply the KL remainder identity in \cref{eq:kl-values}:
\begin{equation}
 L_\beta=\beta\sum_i p_i E_2(s_i-h_i)
 +(1-\beta)\sum_i q_i E_2(-h_i).
 \label{eq:jsd-value}
\end{equation}
Evaluate $E_2$ using \cref{eq:remainder}.
\end{reformulation}

\examplepair
  {\begin{inlinepaperlisting}[listingcomparison]{jsd}{Generalized JSD example code}
\begin{minipage}[t]{.485\linewidth}
\codeheading{issueink}{Basic}
\begin{lstlisting}[style=paperpython,firstnumber=1]
def jsd_basic(z, z_ref, beta):
    p, q, lp, lq, _ = categorical(z, z_ref)
    lm = jnp.logaddexp(
        jnp.log(beta) + lp,
        jnp.log1p(-beta) + lq,
    )
    terms = beta * p * (lp - lm)
    terms += (1 - beta) * q * (lq - lm)
    return jnp.sum(terms)
\end{lstlisting}
\end{minipage}\hfill
\begin{minipage}[t]{.485\linewidth}
\codeheading{clearink}{Fixed}
\begin{lstlisting}[style=paperpython,firstnumber=1]
@custom_gradient
def jsd_fixed(z, z_ref, beta):
    assert_close(z, z_ref)
    p, q, _, _, s = categorical(z, z_ref)
    hq = -jnp.log1p(beta * jnp.expm1(-s))
    terms = (
        beta * p * e2(s - hq)
        + (1 - beta) * q * e2(-hq)
    )
    value = jnp.sum(terms)
    centered = hq - jnp.sum(q * hq)
    gradient = (1 - beta) * q * centered
    return value, gradient
\end{lstlisting}
\end{minipage}
\end{inlinepaperlisting}
}
  {\centering
\includegraphics[width=1.0\linewidth,trim=0 4pt 0 0,clip]{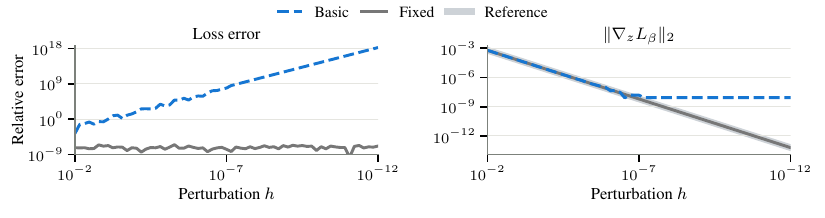}
\papercaption[figure]{Rounding mixture log-ratios can make JSD gradients inaccurate.}{This example uses mixture weight $\beta=1/2$, fixed reference logits $z_{\mathrm{ref}}=(-1,0,1)$ and varying logits $z=(-1,h,1)$.}
\label{fig:example-jsd}
}

\textbf{Gradient computation.}
The same mixture log-ratio determines the gradient. Differentiating
both KL terms in \cref{eq:jsd-loss}, including the dependence of $m$ on
$q$, gives $\nabla_q L_\beta=(1-\beta)h$. The softmax Jacobian then
centers $h$ under $q$, as it centers $s$ for reverse KL:
\begin{equation}
 \nabla_z L_\beta=(1-\beta)q\Bigl(h-\sum_iq_i h_i\Bigr)
 \quad\text{where}\quad h=\log q-\log m.
 \label{eq:jsd-grad}
\end{equation}
Evaluating $h$ through \cref{eq:jsd-fix} avoids the log-probability
subtraction and ratio rounding in \cref{issue:jsd}, retaining the small
differences that centering needs.
\begin{reformulation}[Fixes \cref{issue:jsd} for the JSD gradient in \cref{eq:jsd-grad}]
\label{fix:jsd}
Evaluate $h$ with the $\logp$--$\expm$ expression in
\cref{eq:jsd-fix}, then use it in \cref{eq:jsd-grad}.
The gradient formula is unchanged. The correction preserves the small
mixture log-ratio before centering it.
\end{reformulation}

\Cref{lst:jsd} combines the remainder evaluation with the centered gradient rule.
\Cref{fig:example-jsd} compares it with the direct KL sums using a
log-sum-exp mixture.

\section{Related work}
\label{sec:related}
\textbf{Stable evaluation of divergences.}
The reformulations use classical techniques for avoiding cancellation
\citep[Sections~1.7 and~1.18]{higham2002accuracy}, together with stable
evaluations of elementary functions \citep[Chapter~4]{brent2010arithmetic}. Computational divided differencing similarly
transforms function differences to avoid subtracting nearby evaluations
\citep{reps2003divided}.

Some divergence computations also have established numerical treatments.
\citet[Section~II]{raj2008nonnegative} derive a series with nonnegative grouped terms
for small weighted JSD, a direct precedent for the forward analysis in
\cref{sec:jsd}. \citet[Section~2, Eqs.~(16)--(21), and Appendix~A]{maechler2021remainders}
discusses stable evaluation of $x\log(x/M)+M-x$, the scalar extended-KL term.
The note also analyzes the \texttt{log1pmx} implementation of \citet{welinder2005pgamma}
for stably evaluating $\log(1+x)-x$. Using $\logp(x)$ alone still leaves cancellation when
subtracting $x$, just as $\expm(x)-x$ can lose the quadratic remainder
in our KL, JSD, and K3 formulas. The \texttt{expm1x} routine of
\citet[pp.~36--37]{maechler2026dpq} stably evaluates this same exponential remainder
using a Taylor expansion near zero. Our reformulations
use this established strategy to evaluate divergences.

\textbf{Softmax accuracy and training stability.}
\citet{blanchard2021softmax} analyze rounding
errors in softmax and log-sum-exp, which supply the normalization
primitives for the categorical objectives here. Accurate normalization
alone does not ensure an accurate difference between two nearby
probabilities. \citet{barron2026softmax} address this distinction for
soft cross-entropy gradients. The reverse-KL and generalized JSD
evaluations in \cref{sec:reverse,sec:jsd} build on their use of logit
differences. Related numerical issues arise in attention.
\citet[Sections~3--4]{chen2026brokensymmetry} show how BF16 rounding
breaks the zero-sum constraint on softmax score gradients and contributes
to training instability. Their corrections to the query and key gradients
share our use of an exact gradient identity to reduce numerical error.
\citet[Sections~3--4]{maaz2026property} used property-based
tests to find NaN gradients in the softmax code in the first version of
\citet{barron2026softmax}.
We similarly test our reformulations.

\textbf{Monte Carlo gradient estimation.}
Monte Carlo estimators have different bias and variance tradeoffs even
in exact arithmetic.
\citet{tang2025pitfalls} and \citet{shah2025comedy} analyze the expected
updates produced by Monte Carlo estimates to the reverse KL with samples
from the optimized policy. As discussed in
\cref{sec:monte-carlo-kl}, estimator bias can persist in exact arithmetic,
whereas the numerical issues concern evaluating the chosen estimator.
The reformulations retain the expected update without removing bias or
sampling variability.

\section{Conclusion}
\label{sec:conclusion}
The numerical issues studied here arise when categorical distributions
are parameterized by logits and softmax normalization is composed with
divergence evaluation. Analyzing these operations and their gradients
together leads to reformulations that improve accuracy near equality.
This analysis can be extended to other divergences and loss functions, accounting for their
parameterizations and how their operations are grouped or fused.
A growing set of tools can help automatically check the numerics:
property-based tests can expose failure cases \citep{maaz2026property},
while formal verification can prove numerical error bounds
\citep{george2026rounding}.

\subsection*{Acknowledgments}
I would like to thank
Kyle Kastner,
Lucas Maystre,
Oliver Liu,
Risto Vuorio,
and
Sam Buchanan for insightful comments and discussions.
Language models helped with writing, code, and numerical checks.

\bibliographystyle{bda_style}
\bibliography{refs}
\appendix
\section{Numerical setup and helper implementations}
\label{app:helpers}
\textbf{Numerically stable primitives.}
We assume access to the standard routines $\expm(x)$ and $\logp(x)$,
which evaluate $e^x-1$ and $\log(1+x)$ in a numerically stable way.
\citet[Exercises~4.8--4.9]{brent2010arithmetic} describe how series and
range reduction preserve these increments without first rounding
$e^x$ or $1+x$ near one.

\textbf{Experiments and code.}
The paper provides small numerical examples to illustrate the issues
and fixes, comparing basic and fixed FP32 evaluations on a CPU.
Reference values and gradients use \texttt{mpmath} with 80 decimal digits.
Following \citet{maaz2026property}, we use Hypothesis
\citep{maciver2019hypothesis} for property-based testing of all FP32
reformulations against these references. The~categorical checks cover
equality, category permutations, representable common logit shifts,
and forward and gradient accuracy for $K\in\{2,3,4,8,16\}$.

\textbf{Custom gradients.}
The \texttt{custom\_gradient} decorator wraps a function returning
\texttt{(value, gradient)}, with the derivative taken with respect to its
first input by default. The decorated function returns only the value and
uses the supplied derivative for backpropagation.
For K3 (\cref{lst:k3}), \texttt{argnums=(0, 1)} selects $z_a$ and $\log Z$,
with derivatives returned in that order as \texttt{(g, -g)}.
The remaining inputs are held fixed.

\textbf{Log-ratio reconstruction.}
The helper \texttt{categorical} reconstructs
the log-ratio in \cref{eq:logratio} from logit differences and returns
$p,q,\log p,\log q,s$.

\begin{listingframe}[helperlisting]
\begin{lstlisting}[style=paperpython,firstnumber=1]
def categorical(z, z_ref):
    lp = jax.nn.log_softmax(z_ref)
    lq = jax.nn.log_softmax(z)
    p, q = jnp.exp(lp), jnp.exp(lq)
    delta = z - z_ref
    delta = delta - delta[0]
    d = delta - jnp.sum(p * delta)
    s = d - jnp.log1p(jnp.sum(p * jnp.expm1(d)))
    return p, q, lp, lq, s
\end{lstlisting}
\end{listingframe}

\textbf{Exponential remainder.}
The helper \texttt{e2} evaluates
$E_2(x)=e^x-1-x$ using the series in \cref{eq:remainder} for $|x|<1/2$
and $\expm(x)-x$ otherwise. Its degree-12 polynomial retains the
quadratic leading term without subtracting first-order quantities.
\citet[\texttt{expm1x}, pp.~36--37]{maechler2026dpq} uses the same Taylor
strategy. Exponential remainders also appear in exponential integrators
\citep[Eqs.~(2.10)--(2.11)]{hochbruck2010exponential}. Ordinary autodiff of the polynomial retains
the small derivative in the plotted examples. An explicit backward rule
can instead use $E_2'(x)=\expm(x)$, as in \cref{lst:k3}.

\begin{listingframe}[helperlisting]
\begin{lstlisting}[style=paperpython,firstnumber=1]
def e2(x, degree=12, cutoff=0.5):
    poly = jnp.full_like(x, 1 / math.factorial(degree))
    for k in range(degree - 1, 1, -1):
        coeff = 1 / math.factorial(k)
        poly = coeff + x * poly
    return jnp.where(jnp.abs(x) < cutoff, x * x * poly, jnp.expm1(x) - x)
\end{lstlisting}
\end{listingframe}

\textbf{Closeness guard.}
The full-distribution KL and JSD implementations call \texttt{assert\_close}
before reconstructing the log-ratio.
It checks that the difference between the largest and smallest entries of
$\delta=z-z_{\mathrm{ref}}$ is finite and at most \texttt{limit}, which
defaults to $1/2$. In exact arithmetic, this bounds every $|s_i|$ by the same limit:
$\log\bigl(\sum_i p_i e^{\delta_i}\bigr)$ lies between the smallest and
largest $\delta_i$. The condition is invariant to common logit shifts in
exact arithmetic. The implemented check uses the rounded logit differences
and controls the exponential range without guaranteeing numerical accuracy.
The listing uses an ordinary assertion. The compiled JAX examples check
the same condition with \texttt{checkify}.

In the rounding example of \cref{fig:example-logit_rounding}, the exact
logit differences have a span of $h=2^{-25}$ and the rounded differences $(1,1)$ have
a span of zero, so both pass the guard. Both FP32 softmax outputs also equal
$(1/2,1/2)$, so checking the probabilities would not detect this failure.

\begin{listingframe}[helperlisting]
\begin{lstlisting}[style=paperpython,firstnumber=1]
def assert_close(z, z_ref, limit=0.5):
    delta = z - z_ref
    span = jnp.max(delta) - jnp.min(delta)
    assert jnp.isfinite(span) & (span <= limit)
\end{lstlisting}
\end{listingframe}

\end{document}